\documentclass[10pt,twocolumn,letterpaper]{article}
\usepackage{authblk}
\usepackage[pagenumbers]{cvpr}
\usepackage{times}
\usepackage{graphicx}
\usepackage{epsfig}

\usepackage{multirow,tabularx}
\usepackage[point]{fltpoint}
\usepackage{amsmath,amssymb}
\usepackage{ifthen}
\usepackage{color}
\usepackage{xcolor}
\usepackage{tabularx}
\usepackage{booktabs}
\usepackage{afterpage}
\usepackage{float}
\usepackage{lipsum}

\usepackage[listofformat=parens, justification=centering, subrefformat=subparens]{subfig}

\usepackage[pagebackref=true,breaklinks=true,colorlinks,bookmarks=false]{hyperref}

\renewcommand\Affilfont{\itshape\small}
\makeatletter
\renewcommand\AB@affilsepx{, \protect\Affilfont}
\makeatother

\DeclareGraphicsExtensions{.pdf,.jpg,.png}
\graphicspath{{./figures/}}

\newcolumntype{C}{X<{\centering}}

\definecolor{lightgreen}{rgb}{0.67, 0.88, 0.69}
\definecolor{darkgreen}{rgb}{0,0.55,0}
\definecolor{linkcolor}{rgb}{0,0,.65}

\newcommand\Caption[3][]{\caption[#2]{\label{#1}\textsc{#2}. \small\it#3}}

\DeclareMathOperator*{\argmax}{\arg\,\max}

\newcommand\fig[1]{Fig.~\ref{fig:#1}}
\newcommand\sfig[1]{Fig.~\subref{fig:#1}}
\newcommand\tab[1]{Tab.~\ref{tab:#1}}
\newcommand\stab[1]{Tab.~\subref{tab:#1}}

\usepackage{fancyhdr}

\begin{document}

\title{Watchlist Challenge: 3\textsuperscript{rd} Open-set Face Detection and Identification}

\author[a]{F. Kasım}
\author[b]{T. E. Boult}
\author[c]{R. Mora}
\author[d]{B. Biesseck}
\author[e]{R. Ribeiro}
\author[f]{J. Schlueter}
\author[g]{\\T. Repák}
\author[c]{R. Vareto}
\author[d]{D. Menotti}
\author[c]{W. R. Schwartz}
\author[a]{M. Günther}

\affil[a]{University of Zurich}
\affil[b]{University of Colorado Colorado Springs}
\affil[c]{Federal University of Minas Gerais}
\affil[d]{Federal University of Paraná}
\affil[e]{Federal Police of Brazil}
\affil[f]{DERMALOG Identification Systems GmbH}


\maketitle
\fancyhead[C]{\small This is a pre-print of the original paper accepted for presentation at the International Joint Conference on Biometrics (IJCB) 2024.}
\thispagestyle{fancy}

\begin{abstract}
In the current landscape of biometrics and surveillance, the ability to accurately recognize faces in uncontrolled settings is paramount. 
The Watchlist Challenge addresses this critical need by focusing on face detection and open-set identification in real-world surveillance scenarios. 
This paper presents a comprehensive evaluation of participating algorithms, using the enhanced UnConstrained College Students (UCCS) dataset with new evaluation protocols. 
In total, four participants submitted four face detection and nine open-set face recognition systems.
The evaluation demonstrates that while detection capabilities are generally robust, closed-set identification performance varies significantly, with models pre-trained on large-scale datasets showing superior performance. 
However, open-set scenarios require further improvement, especially at higher true positive identification rates, \ie, lower thresholds.
\end{abstract}

\section{Introduction}

In the evolving field of biometrics, face recognition technology stands as a cornerstone for security and surveillance systems worldwide. 
Surveillance systems, particularly in uncontrolled spaces, frequently encounter significant challenges such as blurry, partially occluded, or poorly illuminated facial images.
In real-world use, the effectiveness of face recognition systems hinges on their ability to perform well under these tough conditions, which starkly differ from the controlled settings that most research focuses on and that often feature high-quality, cooperative subjects, typically containing only celebrities.
Furthermore, the unpredictable nature of real-world surveillance necessitates a departure from traditional closed-set environments, where the gallery of subjects is identical to those employed for probing. 
Instead, the focus shifts to the watchlist problem, where the objective is to identify a few individuals listed on a watchlist while ignoring others (unknowns) \cite{guenther2020watchlist}. 
Such watchlists can be used for identifying missing people, prohibition of unauthorized entry, or capturing criminals.
Especially the latter scenario raises significant risks, including the wrongful targeting of innocent people \cite{hill2020wrongfully}, increased operational costs, and potential liabilities for security staff. 
These challenges underscore the critical need for solutions capable of reliable operation in open-set environments.

\begin{table}
  \small
  \begin{center}
  \begin{tabular}{|l|c|c|c|c|}
    \hline
    \textbf{Sets} & \footnotesize\textbf{Images} & \footnotesize\textbf{Faces} & \footnotesize\textbf{Known} & \footnotesize\textbf{Unknown} \\
    \hline
    Watchlist  & --- & 10000 & 10000 & --- \\
    \hline
    Validation & 7584 & 17689 & 9396 & 8293 \\
    \hline
    Test & 20534 & 57368 & 31512 & 25856 \\
    \hline
  \end{tabular}
  \end{center}
    \Caption[tab:distribution]{Label Distribution}{Distribution of images, faces, known and unknown labels are provided for all sets. The watchlist comprises cropped face regions corresponding to 1000 distinct identities, as shown in \sfig{UCCS:watchlist}.}
\end{table}

\begin{figure*}
\begin{center}
    \subfloat[Example Images\label{fig:UCCS:images}]{
        \begin{tabular}{cc}
            \includegraphics[width=.4\textwidth]{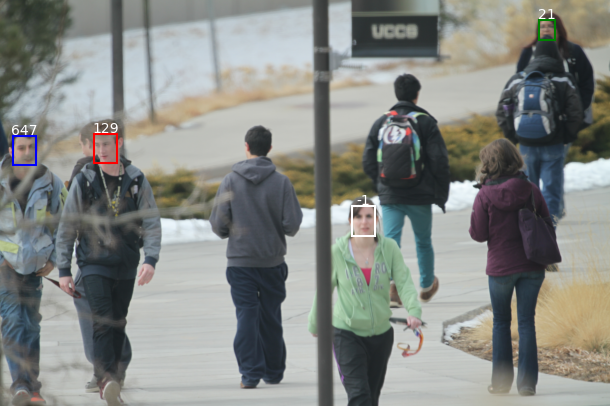} & 
            \includegraphics[width=.4\textwidth]{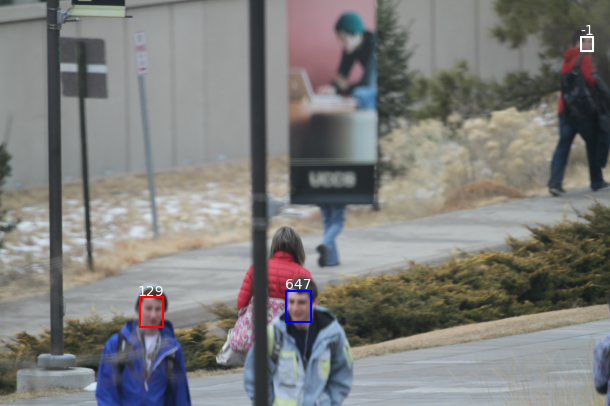}
        \end{tabular}
    }
    \hspace{0.001mm} 
    \subfloat[Watchlist\label{fig:UCCS:watchlist}]{
    \begin{tabular}{c}
            \includegraphics[width=.10\textwidth]{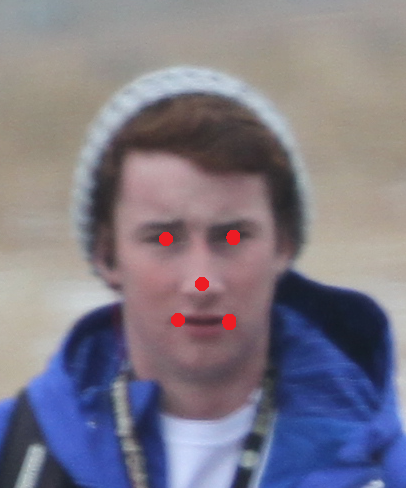} \\
            \includegraphics[width=.10\textwidth]{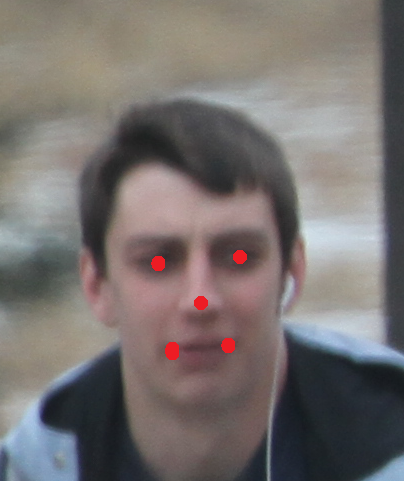}
        \end{tabular}
    }
\end{center}
    \Caption[fig:UCCS]{Example Images and Watchlist}{\subref*{fig:UCCS:images} shows two images with their annotations from the new version of UCCS dataset, including occlusions, different angles, and instances of significant blur. Faces marked with the same color indicate the same identity, whereas white boxes denote unknown subjects. \subref*{fig:UCCS:watchlist} displays cropped faces in the watchlist, including 5 facial landmarks.}
\end{figure*}

In our challenge involving face detection and open-set identification, we exploit the UnConstrained College Students (UCCS) dataset introduced by Sapkota and Boult in 2013~\cite{sapkota2013large}, and substantially increased in size \cite{guenther2017challenge,guenther2018challengeposter} and label quality (see Supplemental Material).
Specifically, we responded to previous criticism\footnote{\url{https://exposing.ai/uccs}} by re-encoding the images to remove EXIF file information revealing details on the dataset collection, which itself was covered by IRB approval.
This dataset is particularly well-suited to reflecting the surveillance system challenges discussed earlier, making it an ideal tool for addressing the critical issues encountered in real-world settings for the watchlist problem.
In this dataset, individuals are typically unaware that their images are being recorded, which mirrors the non-cooperative behavior found in live surveillance scenarios and adds to the complexity of identifying watchlist subjects.
One unique property of this dataset is that images are captured across different weather conditions, including sun with strong cast shadows, but also rainy and snowy conditions that severely influence imaging conditions.
Additionally, the challenge incorporates face detection tasks where detectors may erroneously select regions of the background as faces. 
Along with a large pool of faces of unknown/innocent subjects that should not induce false identifications, these background detections are crucial to address, as they must also be treated as unknown by face recognition algorithms to provide a complete benchmark for evaluation.
By focusing on these aspects, our challenge aims to foster research and development in face detection and recognition technologies, which are increasingly vital in today's surveillance applications.

This challenge has been conducted twice in previous research \cite{guenther2017challenge,guenther2018challengeposter}.
Both instances yielded satisfactory outcomes in evaluating the detection capabilities and closed-set recognition of algorithms.
However, open-set face recognition, which entails recognizing unknown faces and dealing with misdetections, remains a significant unresolved challenge.
This complexity is particularly evident when probe faces are captured on different days than the gallery. 
Additionally, previous evaluation protocols displayed inherent biases as both enrollment and probe data were randomly selected from the same set, leading to unrealistic \emph{same condition} matches.
Past competitions also revealed numerous inaccuracies in the labels of some identities. 
To address these issues, the evaluation protocol has been revised to minimize the temporal overlap between gallery and probe data, especially in the test set. 
Simultaneously, the dataset underwent a significant cleanup, employing a combination of semi-automated and manual methods.

The UCCS Watchlist Challenge\footnote{\url{https://www.ifi.uzh.ch/en/aiml/challenge.html}} is structured into two distinct segments: (I) face detection and (II) open-set face recognition.
Participants contributed by submitting their results for these specialized tasks. 
In Part I, the challenge is to detect all faces within the UCCS images, irrespective of the identity labels, ensuring comprehensive coverage of face detection capabilities.
Part II requires participants to enroll a set of gallery identities and then compute the similarities between each detected face (including various false positive detections) from Part I and each identity in the watchlist. 
These similarity scores are critical, as they determine which faces match the watchlist identities, while effectively ignoring unknown/unimportant faces.

\section{Dataset and Protocol}
\label{sec:database}
The UCCS dataset was collected over several months on the private premises of the University of Colorado Colorado Springs.
Two exemplary images from the UCCS dataset are presented in \sfig{UCCS:images}, illustrating the diversity in facial orientations, occlusions, and blur.
In response to previously identified issues with incorrectly labeled identities \cite{guenther2017challenge,guenther2018challengeposter}, a systematic data cleaning process was implemented, straightening detection, missing labels, as well as intra-class and inter-class label issues. 
Further details on the original UCCS dataset and these cleaning procedures can be found in the Supplemental Material, as well as a comparison to related surveillance datasets SCface~\cite{grgic2011scface}, PaSC~\cite{PaSC}, IJB-S~\cite{ijbsdataset}, DroneSurf~\cite{dronesurfdataset} and BRIAR~\cite{briardataset}.

\subsection{Defining new Evaluation Protocols}

In contrast to prior invocations of this competition \cite{guenther2017challenge,guenther2018challengeposter}, this year's challenge introduces a separate watchlist, which comprises cropped expanded face regions from the dataset, eliminating the need for a traditional training set. 
Ten faces per watchlist identity were extracted from the images, and subsequently excluded from part II of our evaluation.
Similarly to previous encounters of this competition, we did not include all labeled identities into our watchlist, but left some of them to be unknown.
The selection of the watchlist prioritized identities that contain high-quality faces and appear in multiple sequences, while 10 gallery faces were extracted from the same sequence only.
These annotations of the selected faces include information on the positions of five automatically detected  facial landmarks, including the eyes, nose, and mouth.
Following these constraints, a selection process resulted in choosing precisely 1000 different identities that are treated as known subjects in our watchlist.
About 40\% of these identities appear over two or more days. 
\sfig{UCCS:watchlist} illustrates different gallery faces and their annotations originating from two identities.

The final version of the UCCS dataset consists of more than 85'000 faces in total. 
The distribution of labeled faces in our dataset can be found in \tab{distribution}.
Following the completion of the gallery, we split up the images into validation and test sets.
The validation set includes annotated images with lists of bounding boxes, each labeled either with an integral gallery identity label or with the unknown label $-1$.
In contrast, the test set consists solely of raw images with anonymized filenames and no annotations, challenging participants to detect faces and provide similarity scores for each bounding box against all watchlist subjects.
To enhance realism and reduce the potential for biased \emph{same day} matches, the test set includes images from different sequences of the watchlist identities, if available, whereas the validation set predominantly uses faces from the same sequences.
Ultimately, 996 known identities appear in the test set, compared to 932 in the validation set. 
In both sets, about half of the faces are categorized as \emph{unknown} to emulate real-world scenarios, including the remaining subjects that are not enrolled in the gallery, and faces that are left with the unknown label in our dataset.

\section{Challenge Participants}
\label{sec:participants}

Participants were invited to contribute summaries of their algorithms. 
Together with baseline algorithms, they are presented in the order of submission and tagged with their respective institutions ({\textsuperscript{\emph{a}} -- \textsuperscript{\emph{g}}}, \cf~list of authors).

\subsection{Face Detection}
\label{sec:participants:detection}
\textbf{MTCNN-Baseline}: The baseline face detector simply uses the pre-trained MTCNN \cite{zhang2016mtcnn}, with its Pytorch implementation.\footnote{\url{https://pypi.org/project/facenet-pytorch}} 
Since the detector is not optimized for blurry, occluded, or full-profile faces, we had to lower the three detection thresholds to ($0.2$, $0.2$, $0.2$), which ended up in detecting most faces, but provides many false positive detections.
Our implementation can be downloaded from PyPI.\footnote{\label{baselinepack}\url{https://pypi.org/project/challenge.uccs}}

\textbf{RetinaFace}\textsuperscript{\textit{c,d,e}}: The multi-task face detector \cite{retinaface_Deng2020} is designed to identify face bounding boxes and five key facial landmarks. 
This detector employs a feature pyramid and anchor boxes in its pipeline. 
It is trained on the WIDER FACE dataset \cite{yang2016wider} using a multi-task loss function.
The model was used at multiple scales with image factor sizes of $(0.2, 0.5, 1.0)$, a confidence threshold of $0.3$, and a Non-Maximum Suppression (NMS) IoU of $0.075$. 

\textbf{F3Y640S/F3Y640L}\textsuperscript{\textit{f}}: The DERMALOG Face SDK\footnote{\label{dermalog}\fontsize{6}{8}\selectfont\url{https://www.dermalog.com/products/software/face-recognition}} implements the entire proprietary facial recognition system, including face detection, landmark detection, and feature extraction. 
The F3Y640S and F3Y640L face detection models are based on the YOLOX architecture \cite{yolox2021}.
This single-shot architecture leverages spatial pyramid pooling, a feature pyramid, and decoupled heads.
The F3Y640S model is less complex, featuring reduced network depth and fewer parameters compared to the F3Y640L.
Face detections by these models are filtered based on the confidence scores they generate.
Training for these systems has been performed using both publicly available and commercially usable datasets, along with additional internal data.

\subsection{Face Identification}

\textbf{MagFace-Baseline}: The baseline feature extractor\footnote{\url{https://github.com/IrvingMeng/MagFace}} utilizes the MagFace model \cite{meng2021magface}, which is pre-trained on the MS1MV2 dataset \cite{guo2016ms,deng2019arcface} using an iResNet-100 backbone \cite{duta_improved_2021}.
MagFace aims to increase the inter-class distance while maintaining a cone-like structure within each class to ensure ambiguous samples are pushed toward the origin and away from class centers.
Faces detected by the baseline detector are aligned based on facial landmarks \cite{meng2021magface}.
Each face is represented by a 512-dimensional embedding.
The enrollment averages embeddings from 10 faces per subject.
Probe faces are compared to the gallery via cosine similarity.

\textbf{AdaFace}\textsuperscript{\textit{c,d,e}}: ResNet-100 \cite{he2016deep} is trained on the MS1MV3 \cite{guo2016ms} dataset with the AdaFace \cite{kim2022adaface} loss function, which introduces an angular margin loss that utilizes image quality to scale the gradient during training and adjusts margins for different classes based on their recognition difficulty.
Image quality is assessed using the norm of the 512-dimensinoal embedding. 
The process for enrollment and probing aligns with the baseline method.

\textbf{MEL}\textsuperscript{\textit{c,d,e}}: This method \cite{vareto2024open} integrates the principles of Maximal Entropy and Objectosphere Loss \cite{dhamija2019improving} to enhance face recognition capabilities.
Maximal Entropy increases the entropy in feature representations, effectively reducing certainty about any unknown category to distinguish more accurately between known and unknown faces.
Concurrently, Objectosphere loss improves the separation of known and unknown classes within the feature space. 
To train this MEL model, it receives the AdaFace embeddings as inputs, projects them to a new space where the unknown class is more compacted and separated from known person classes.
This training includes the known gallery embeddings and an unknown class that is composed of 1600 unknown subjects extracted from the UCCS validation set.
Enrollment and scoring follow the baseline protocol.

\textbf{F3Y640S/F3Y640L}\textsuperscript{\textit{f}}: Before feature extraction, a novel facial landmark detection\footref{dermalog} assesses face quality by evaluating the occlusion of each landmark, setting thresholds to exclude faces that could yield poor matches.
Features are then extracted using the same face recognition model\footref{dermalog} for faces detected separately by the F3Y640S and F3Y640L detectors.
All enrollment faces for each identity are encoded into templates
and used during the matching process.
The query template (probe face) is compared against the templates of all enrolled identities.
Only the highest matching score achieved between the query template and each enrolled identity’s templates is reported in the score file.

\textbf{DaliFace}\textsuperscript{\textit{a,b}}: DaliFace \cite{robbins2023daliid} uses ResNet-100 \cite{he2016deep} as its backbone, trained on the WebFace4M \cite{zhu2021webface260m} dataset with the AdaFace \cite{kim2022adaface} loss function.
It incorporates distortion augmentations during its training to simulate real-world distortions like motion blur and atmospheric turbulence, maintaining feature-level invariance against severe image quality degradations.
Additionally, an adaptive weighting schedule adjusts the intensity of these distortions progressively, allowing the model to adapt gradually without overwhelming initial learning stages.
DaliFace processes the detection results from JointFaceDetectID, which provides only bounding boxes; therefore, the faces are resized for inference.
Enrollment and scoring follows the baseline.

\textbf{EnsembET/EnsembETMN/ET}\textsuperscript{\textit{g}}: EnsembET is an ensemble model composed of three identical EVA-02-Ti transformer-based visual representation models \cite{fang2023eva}, each utilizing a different loss function or training strategy.
All models are trained on facial images resized to $224\times224$, sourced from the LFW \cite{huang2008labeled}, CelebA \cite{liu2015deep}, and the UCCS validation set.
The first model in the ensemble is trained from scratch with triplet loss \cite{Schroff_2015_CVPR} using Euclidean distance, the second with cosine distance. 
The third starts with pre-trained weights and is fine-tuned using triplet loss with cosine distance.
Similarly, the EnsembETMN model is an ensemble setup that changes only the first model with MobileNetV2 \cite{sandler2018mobilenetv2}, while keeping the same training strategies and loss functions as EnsembET.
Finally, the ET model is a single instance of EVA-02-Ti that uses pre-trained weights and is fine-tuned with triplet loss using cosine distance.
The enrollment for the watchlist is similar to the baseline. 
For ensemble models, the final scores are derived by averaging scores from individual components of the ensemble. 

\subsection{Joint Face Detection and Identification}

\textbf{JointFaceDetectID}\textsuperscript{\textit{a,b}}: The JointFaceDetectID model represents a novel approach that integrates face detection and recognition tasks into a single model.
This unified approach uses the same feature space for both tasks, potentially improving accuracy by leveraging their interdependencies.
Designed as a single-stage, anchor-based detector, the model's architecture includes iResNet-50 \cite{duta_improved_2021} backbone, a feature pyramid network \cite{lin2017feature} neck, and heads that consist of three branches.
Two detection branches handle anchor classification and provide bounding boxes, employing focal loss \cite{Lin_2017_ICCV} and distance-IoU loss \cite{zheng2020distance}.
The last branch generates 256-dimensional embeddings to represent faces, utilizing ArcFace loss \cite{deng2019arcface} to enhance the discriminative capabilities of the embeddings.
Furthermore, Entropic Open-Set loss \cite{dhamija2018reducing} is used to manage the uncertainty associated with unknown faces and high-confidence false postive detections.
The model was initially trained on the IJB-C \cite{maze2018iarpa} and PaSC \cite{PaSC} datasets, which collectively contain 3966 identities, before being fine-tuned on the UCCS gallery and validation sets, using faces labeled as $-1$ as negatives. 

During inference, the model generates multiple anchors for each face. 
To finalize the embedding, a confidence threshold of 0.5 filters out low-confidence boxes, and NMS with a 0.4 IoU threshold removes highly overlapping boxes. 
This process ensures that the remaining bounding box accurately represents the face.
The enrollment and scoring processes follow the same methodology as the baseline.

\section{Evaluation}
\label{sec:eval}
\begin{figure*}[t!]
    \centering
    \subfloat[Detection on Validation Set\label{fig:FROC:validation}]{\includegraphics[width=.4\textwidth]{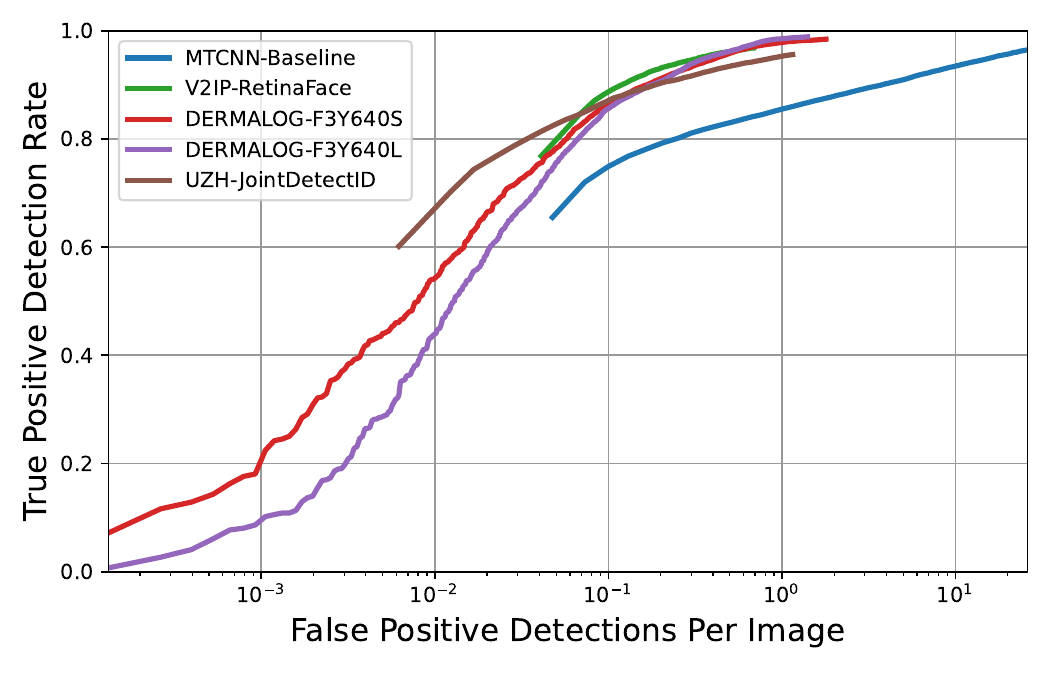}}
    \qquad
    \subfloat[Detection on Test Set\label{fig:FROC:test}]{\includegraphics[width=.4\textwidth]{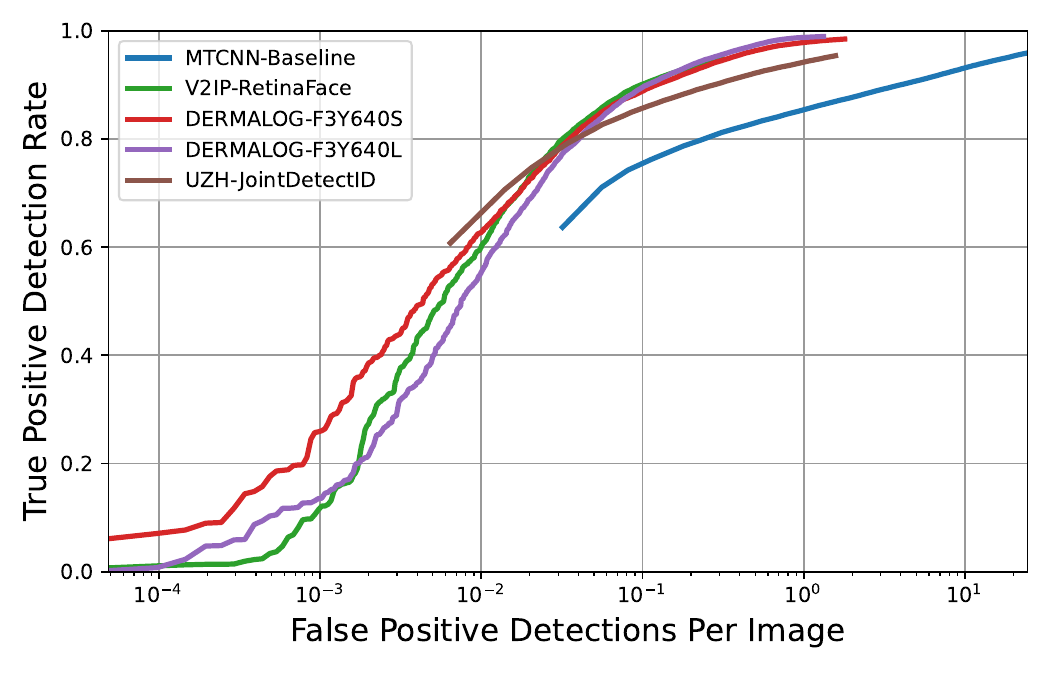}}

    \subfloat[Identification on Validation Set\label{fig:OROC:validation}]{\includegraphics[width=.4\textwidth]{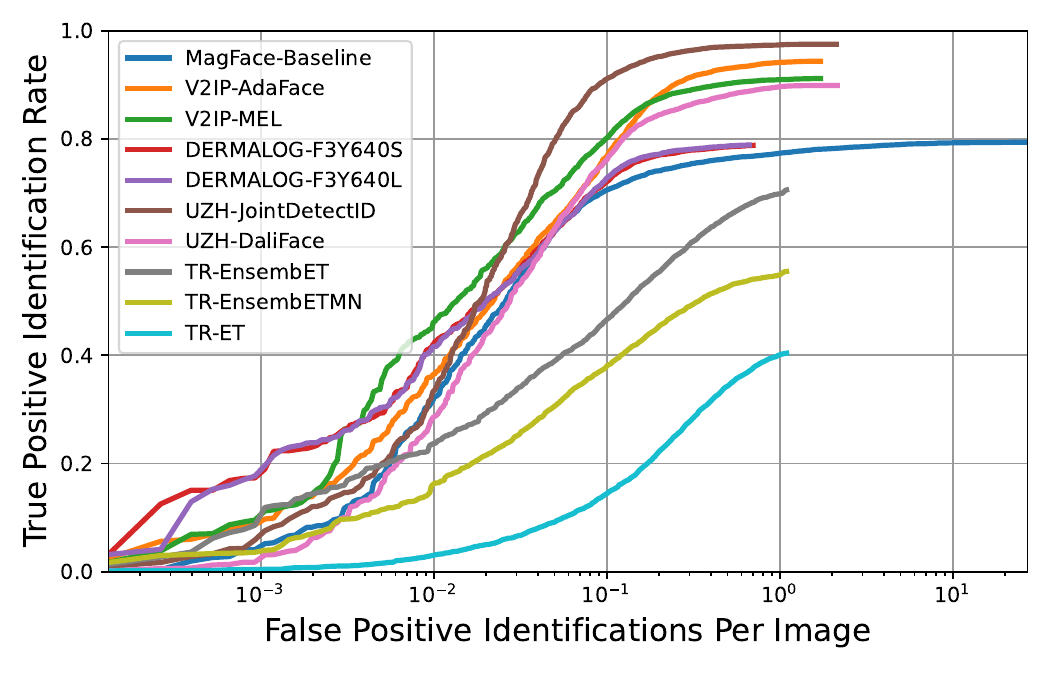}}
    \qquad
    \subfloat[Identification on Test Set\label{fig:OROC:test}]{\includegraphics[width=.4\textwidth]{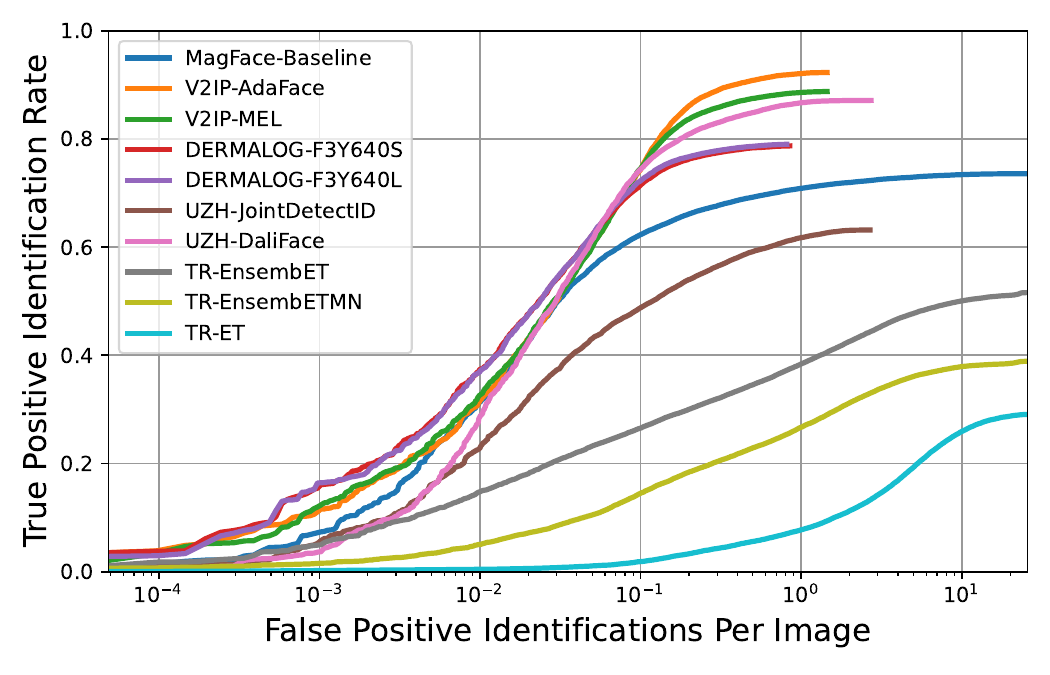}}

    \Caption[fig:FROC]{Face Detection and Recognition Evaluation}{
        A Free-response Receiver Operating Characteristic (FROC) curve is shown for the \subref*{fig:FROC:validation} validation and \subref*{fig:FROC:test} test set. 
        The horizontal axis includes the number of false positive detections normalized by the number of images, while the vertical axis outlines the relative number of true positive detections of faces.
        Open-set ROC curve at rank 1 is shown for \subref*{fig:OROC:validation} validation and \subref*{fig:OROC:test} test set. 
        The horizontal axis includes the number of false positive identifications normalized by the number of images, while the vertical axis outlines the relative number of correctly identified faces.    
    }
\end{figure*}

For evaluating face detection participants submitted bounding boxes for detected faces, each accompanied by a confidence score. 
For face recognition, participants also provide a similarity score for each watchlist subject associated with the detected faces. 
Since the faces on the watchlist are cropped from the original images, those are omitted from the evaluation process.

Participants were provided with the challenge's evaluation scripts and ground-truths\footref{baselinepack} to facilitate the evaluation on the validation set. 
Here, we use the exact same evaluation framework on the test set.

\subsection{Face Detection}

To assess the accuracy of each bounding box, the standard Jaccard index, also known as the Intersection Over Union (IoU), is used to compare the detected bounding box with the ground truth.
In our evaluation, we accept all bounding boxes with an IoU threshold, $IoU \geq 0.2$.
This threshold is selected to compensate for potential inaccuracies\footnote{\label{fn:bbxs}There were a few faces for which the landmark detector failed in the annotation process -- for these we kept the original bounding boxes \cite{guenther2017challenge}.} in the ground truth boxes, allowing for the inclusion of loosely matched detections.

Face detection evaluation is conducted using an adaptation of the Free-response Receiver Operating Characteristic (FROC) curve \cite{chakraborty2002statistical}.
Specifically, confidence scores $c$ are classified as $C^+$ when $IoU \geq 0.2$ and as $C^-$ when $IoU < 0.2$, based on the overlap between the ground-truth and detected bounding boxes.
The True Positive Detection Rate (TPDR) is calculated using labeled faces \textbf{M} from probe samples if the confidence score exceeds the operating threshold $\theta$:
\begin{equation}
    \text{TPDR}(\theta) = \frac1{|\mathbf M|}\,\Bigl|\bigl\{c \mid c \in C^+ \wedge c \geq \theta\bigr\}\Bigr|
\end{equation}
Instead of calculating False Positive Detection Rates on the horizontal axis, this method measures the number of False Positive Detections Per Image (FPDPI) \cite{guenther2018challengeposter}, calculated by dividing total false positive detections by \emph{the number of probe images} \textbf{I}. 
This approach accounts for the varying number of false positive detections that a face detector might produce across different images. 
A false positive detection occurs when the confidence score of a misdetection is larger than threshold $\theta$:
\begin{equation}
    \text{FPDPI}(\theta) = \frac1{|\mathbf I|}\,\Bigl|\bigl\{c \mid c \in C^- \wedge c \geq \theta\bigr\}\Bigr|
\end{equation}
By varying the threshold $\theta$, the FROC plots TPDR over FPDPI.
Finally, a single score to compare across algorithms is computed by obtaining the thresholds $\theta_i$ that result in FPDPI $=10^i$:
\begin{equation}
    \label{eq:sum-tpdr}
    \hspace*{-.5em}\sum\text{TPDR}=\hspace*{-2em}\sum\limits_{i\in\{-3, -2, -1, 0\}}\hspace*{-2em}\text{TPDR}(\theta_i)\quad \theta_i=\text{FPDPI}^{-1}(10^i)
\end{equation}
We add 0 when a low FPDPI is reached by no threshold $\theta$.

\sfig{FROC:validation} and \sfig{FROC:test} display the participants' detection results on both sets, while \stab{ranking:detection} provides detailed information about the rankings on the test set, evaluated via \eqref{eq:sum-tpdr}.
All models demonstrate consistent behavior across both the validation and test sets, except for V2IP-RetinaFace.
Notably, V2IP-RetinaFace does not register any TPDR at the lower FDPIs on the validation set, yet it does show performance on the test set. 
This discrepancy could be due to V2IP-RetinaFace being particularly sensitive to the specific characteristics or quality of the data in the test set.
DERMALOG-F3Y640S is the best-performing detector with a $\sum$TPDR score of 2.752. It shows the highest by far TPDR at the FPDPI $10^{-3}$, indicating its capability in environments where minimizing false positive detections is crucial.
Additionally, it maintains consistently high performance across other FPDPI thresholds.
DERMALOG-F3Y640L and V2IP-RetinaFace also perform well, particularly at higher FPDPI thresholds (more lenient conditions), which suggests these models maintain a good balance between detecting existing faces and controlling false positives under less restrictive conditions.
UZH-JointDetectID does not report data for the strictest threshold ($10^{-3}$), which might suggest a limitation in its ability to handle extremely low false positive detections.
However, it performs exceptionally well at the $10^{-2}$ threshold, achieving the best TPDR among all methods, indicating its effectiveness in slightly less stringent conditions.
To reach a similar TPDR as other methods, the MTCNN baseline records a notably higher number of false positive detections likely due to its overly permissive thresholds.

\begin{table}[tb!]
    \Caption[tab:ranking]{Ranking for detection and recognition tasks}{\subref*{tab:ranking:detection} shows the detection ranking, whereas the identification ranking is indicated in \subref*{tab:ranking:recognition}.}
    \centering
    \renewcommand{\arraystretch}{1.2} 
    \subfloat[Detection\label{tab:ranking:detection}]{\resizebox{\columnwidth}{!}{%
        \begin{tabular}{|l|c|c|c|c|c|}
            \hline
            \textbf{Method} & \multicolumn{4}{c|}{\textbf{@FPDPI}} & \multirow{2}{*}{\textbf{\boldmath{$\sum$ TPDR}}} \\
            \cline{2-5}
             & \boldmath{$10^{-3}$} & \boldmath{$10^{-2}$} & \boldmath{$10^{-1}$} & \boldmath{$10^{0}$} & \\
            \hline\hline
            DERMALOG-F3Y640S & \textbf{0.2585} & \emph{0.6271} & 0.8882 & \emph{0.9782} & \textbf{2.752} \\
            DERMALOG-F3Y640L & \emph{0.1350} & 0.5530 & \emph{0.8958} & \textbf{0.9875} & \emph{2.5713} \\
            V2IP-RetinaFace & 0.1142 & 0.5993 & \textbf{0.9011} & 0.9469 & 2.5615 \\
            UZH-JointDetectID & - & \textbf{0.6583} & 0.8556 & 0.9409 & 2.4548 \\
            MTCNN-Baseline & - & - & 0.7424 & 0.8519 & 1.5943 \\
            \hline
        \end{tabular}%
    }}

    \subfloat[Identification\label{tab:ranking:recognition}]{\resizebox{\columnwidth}{!}{%
        \begin{tabular}{|l|c|c|c|c|c|}
            \hline
            \textbf{Method} & \multicolumn{4}{c|}{\textbf{@FPIPI}} & \multirow{2}{*}{\textbf{\boldmath{$\sum$ TPIR}}} \\
            \cline{2-5}
             & \boldmath{$10^{-3}$} & \boldmath{$10^{-2}$} & \boldmath{$10^{-1}$} & \boldmath{$10^{0}$} & \\
            \hline\hline
            V2IP-AdaFace & 0.1106 & 0.3157 & 0.7434 & \textbf{0.9208} & \textbf{2.0905} \\
            V2IP-MEL & 0.1196 & 0.3261 & \textbf{0.7457} & \emph{0.8859} & \textit{2.0773} \\
            DERMALOG-F3Y640L & \textbf{0.1640} & \emph{0.3695} & 0.7226 & 0.7901 & 2.0462 \\
            DERMALOG-F3Y640S & \emph{0.1543} & \textbf{0.3739} & 0.7142 & 0.7871 & 2.0295 \\
            UZH-DaliFace & 0.0361 & 0.2857 & \emph{0.7435} & 0.8668 & 1.9321 \\
            MagFace-Baseline & 0.0721 & 0.3135 & 0.6227 & 0.7086 & 1.7169 \\
            UZH-JointDetectID & 0.0522 & 0.2275 & 0.4880 & 0.6175 & 1.3852 \\
            TR-EnsembET & 0.0490 & 0.1479 & 0.2651 & 0.3840 & 0.8460 \\
            TR-EnsembETMN & 0.0152 & 0.0503 & 0.1449 & 0.2671 & 0.4775 \\
            TR-ET & 0.0019 & 0.0043 & 0.0186 & 0.0774 & 0.1022 \\
            \hline
        \end{tabular}
    }}
\end{table}

\subsection{Face Identification}

For identification, we rely on comparing a gallery template $T_g$ and a probe face $F_p$ via scoring function $s$.
In this open-set context, a face recognition algorithm aims to achieve three objectives: 
First, for a probe face of a known identity, the corresponding gallery template $T_{g^*}$ should show the highest similarity among all templates.
Second, if the probe face belongs to an unknown identity, the similarities to all gallery templates should be low. 
Finally, any false positive \emph{detections} should be treated as unknown.
In our evaluation, we rely on our adaptation \cite{guenther2018challengeposter,guenther2020watchlist} of the Open Receiver Operating Characteristic (O-ROC) curve.
We split the probe faces into a set of \emph{known} faces $\mathbf K$, as well as a set of unknown faces and false positive detections $\mathbf U$.
We plot the True Positive Identification Rate (TPIR) over the False Positive Identifications Per Image (FPIPI):
\begin{gather}
    \label{eq:TPIR}
    \begin{aligned}
        \hspace*{-.7em}\text{TPIR}(\tau) = \frac{1}{|\mathbf{K}|}\,\Bigl| \bigl\{ F_p \in \mathbf{K} \mid \argmax_{g \in G} s(T_g, F_p) = g^*& \\
        \land \; s(T_{g^*}, F_p) \geq \tau \bigr\} \Bigr|&
    \end{aligned}\\
    \label{eq:FPIPI}
    \text{FPIPI}(\tau) = \frac{1}{|\mathbf{I}|} \Big| \Big\{ F_p \in \mathbf{U} \mid \max_{g \in G} s(T_g, F_p) \geq \tau \Big\} \Big|
\end{gather}
Similarly to the face detection evaluation, we also define a single number for defining a ranking across algorithms:
\begin{equation}
    \label{eq:sum-tpir}
    \hspace*{-.5em}\sum\text{TPIR}=\hspace*{-2em}\sum\limits_{i\in\{-3, -2, -1, 0\}}\hspace*{-2em}\text{TPIR}(\tau_i)\quad \tau_i=\text{FPIPI}^{-1}(10^i)
\end{equation}

\sfig{OROC:validation} and \sfig{OROC:test} display the participants' recognition results on both sets, while \stab{ranking:recognition} provides detailed information about the rankings on the test set, evaluated at four different FPIPI levels via \eqref{eq:sum-tpir}.
V2IP-AdaFace stands out as the top performer with the highest overall TPIR, showcasing robust capabilities across varying thresholds of false positive identification.
In particular, this model performs exceptionally well (92\%) at the most lenient threshold, demonstrating its adaptability in environments where higher rates of false identifications are permissible. 
DERMALOG-F3Y640L and DERMALOG-F3Y640S, on the other hand, show superior performance at the strictest thresholds, making them ideal for applications demanding high accuracy with minimal false positive identifications.
Both V2IP-MEL and UZH-DaliFace exhibit commendable performances at higher tolerance levels for false identifications, successfully recognizing 86-88\% (@FPIPI $=1$) of watchlist subjects.
Almost all models display consistent performance between the validation and test sets.
One exception is UZH-JointDetectID, which excels on the validation set but shows a marked decrease in effectiveness on the test set, underlining potential overfitting issues.
Furthermore, TR-EnsembET, TR-EnsembETMN, and TR-ET lag significantly behind the other models at all assessed thresholds, struggling to identify true positives accurately, which is likely caused by small number of identities in their training datasets. 
However, it is noteworthy that among these, the ensemble models (TR-EnsembET and TR-EnsembETMN) demonstrate better performance compared to TR-ET, indicating that the ensemble approach does offer advantages even among the lower-performing models.

\section{Discussion}
Additionally to the main results, we explore key aspects of the facial recognition challenge, including closed-set performance, threshold effects, and handling of unknowns. 
A detailed analysis outlining specific detection and identification failure cases, along with limitations and potential improvements, is available in the Supplemental Material.

\subsection{Closed-Set Performance}
\begin{table}[tb!]
    \Caption[tab:closeset]{Closed-Set Performance}{Closed-set performance of methods is shown for both tasks.}
    \centering
    \small
    \begin{tabularx}{.99\linewidth}{|X@{}||c||c|c|} 
        \hline
        \textbf{Method} & \textbf{TPDR}(\%) & \textbf{TPIR}(\%) & \textbf{FPIPI} \\ 
        \hline\hline
        MagFace-Baseline & \textbf{98.81} & 73.58 & 25.63 \\\hline
        V2IP-AdaFace & \multirow{2}{*}{98.45} & \textbf{92.27} & \multirow{2}{*}{1.449} \\
        V2IP-MEL &  & \emph{88.77} & \\\hline
        DERMALOG-F3Y640S & 83.30 & 78.71 & 0.848 \\
        DERMALOG-F3Y640L & 83.62 & 79.01 & 0.813 \\\hline
        UZH-JointDetectID & \multirow{2}{*}{\emph{98.80}} & 63.17 & 2.684 \\
        UZH-DaliFace & & 87.12 & 2.724 \\\hline
        TR-EnsembET & \multirow{3}{*}{\textbf{98.81}} & 51.59 & \multirow{3}{*}{25.63} \\
        TR-EnsembETMN &  & 38.90 &  \\
        TR-ET & & 29.07 &  \\
        \hline
    \end{tabularx}
\end{table}

\begin{figure*}
    \centering
    \subfloat[Detection\label{fig:thresholds:detection}]{\includegraphics[width=.4\textwidth]{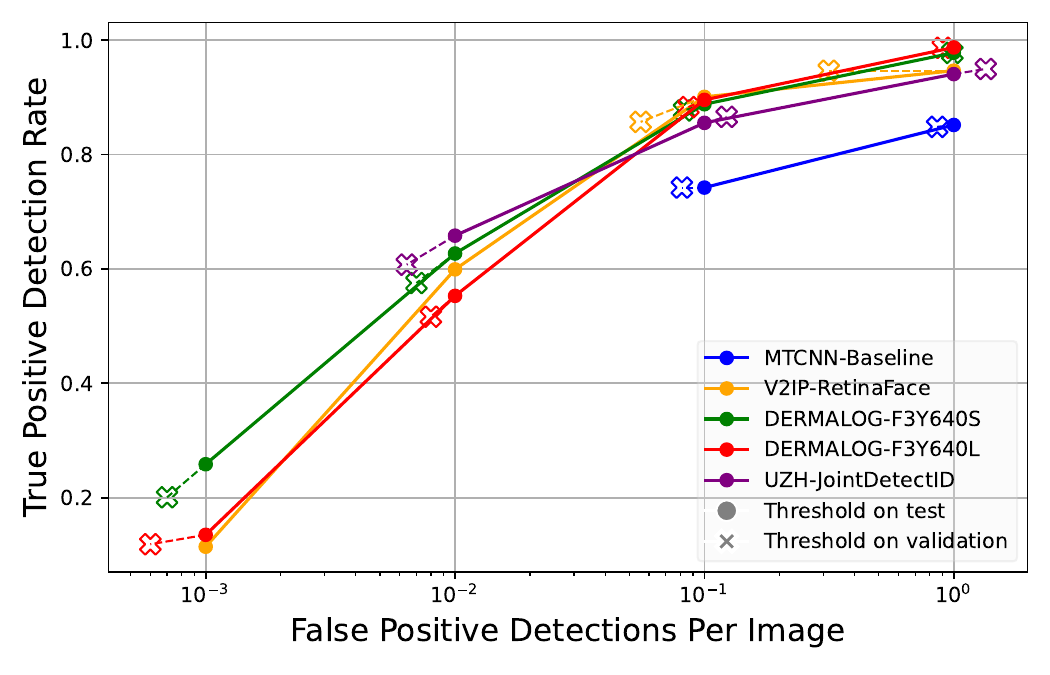}}
    \qquad
    \subfloat[Identification\label{fig:thresholds:recognition}]{\includegraphics[width=.4\textwidth]{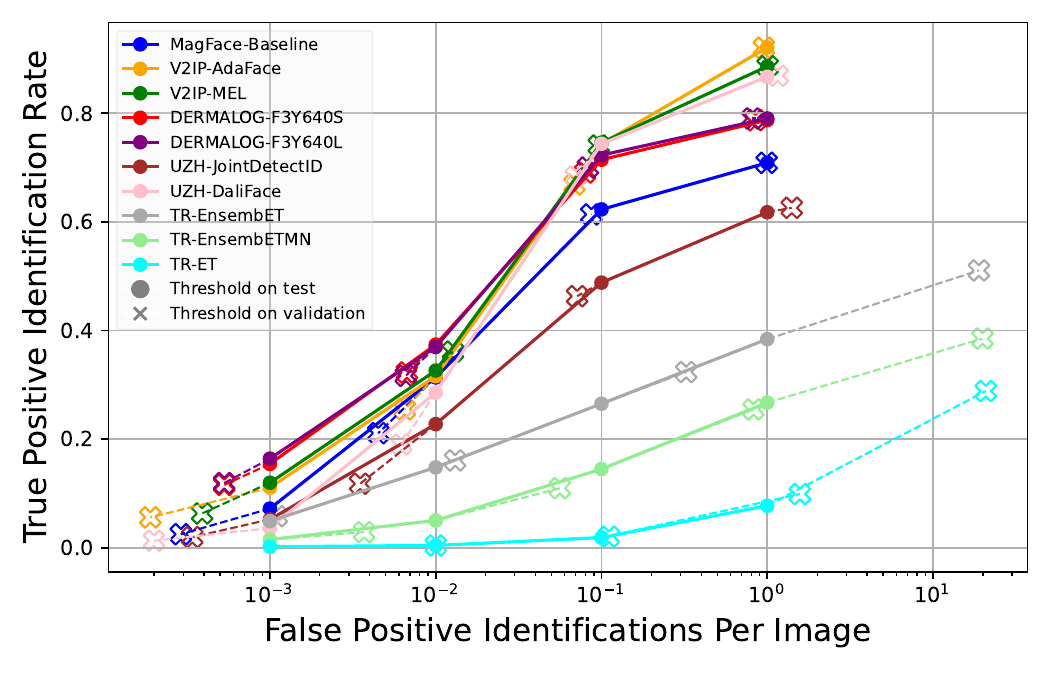}}
    \Caption[fig:thresholds]{Threshold Selection}{
        We depict the effect of selecting the thresholds on the validation and test sets. 
        In \subref*{fig:thresholds:detection}, we show differences in detection scores, while \subref*{fig:thresholds:recognition} highlights differences in identification performances.}
\end{figure*}

\begin{figure*}
    \centering
    \subfloat[Unknown Faces $\mathbf U_{-1}$\label{fig:CCR:unknown}]{\includegraphics[width=.4\textwidth]{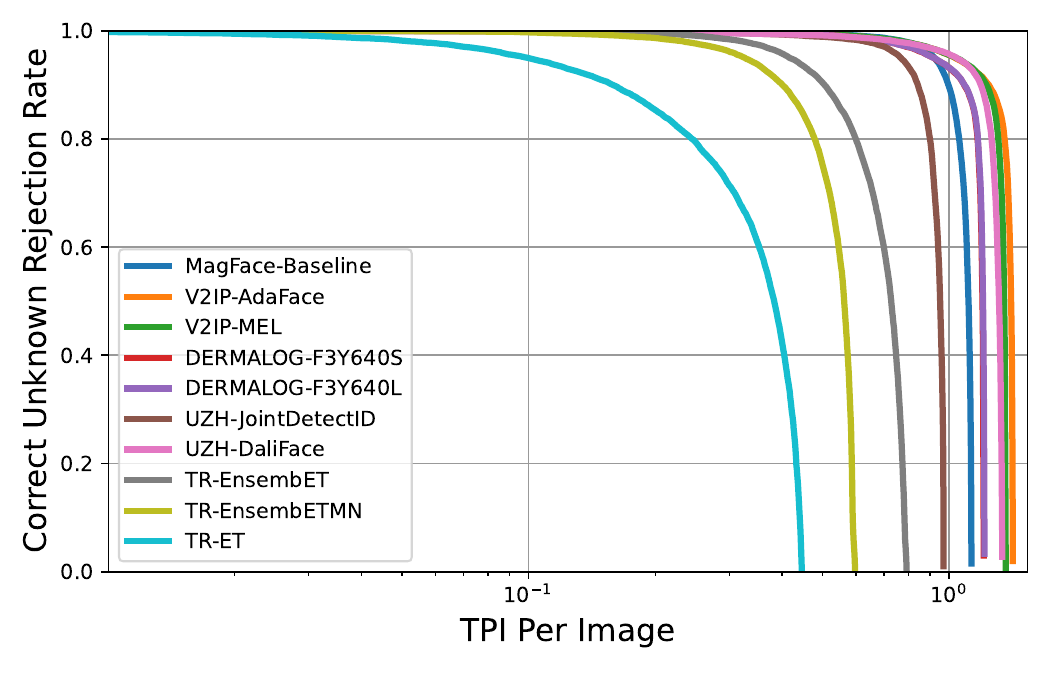}}
    \qquad
    \subfloat[False Positive Detections $\mathbf U_{\mathrm{FPD}}$\label{fig:CCR:background}]{\includegraphics[width=.4\textwidth]{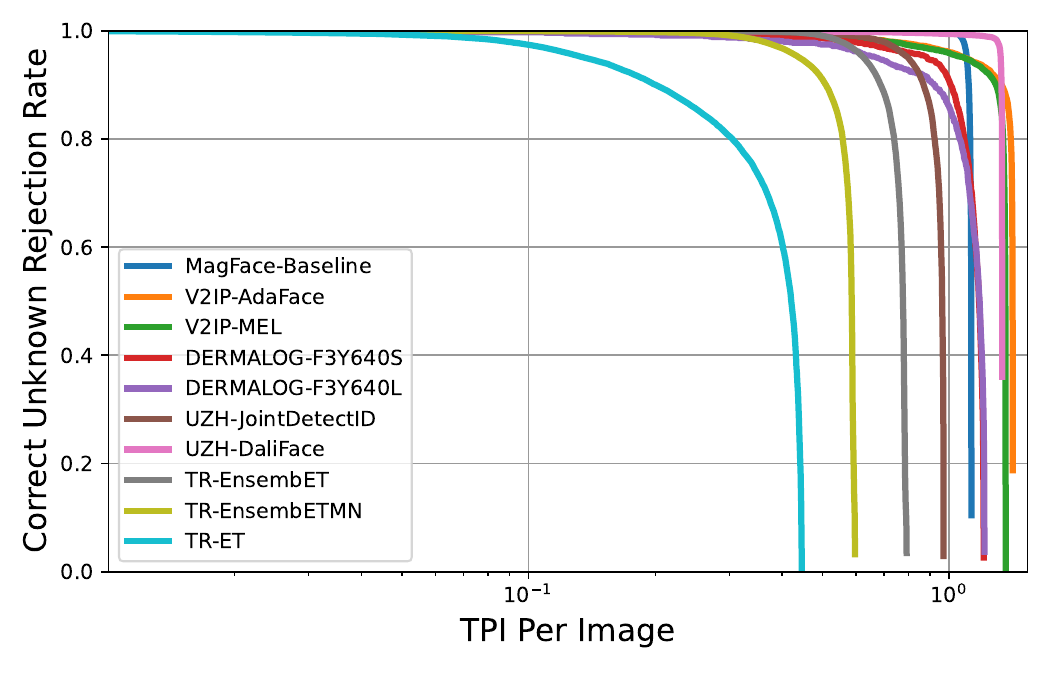}}
    
    \Caption[fig:CCR]{Rejection Rates by Type}{
        Unknown samples are split into \subref*{fig:CCR:unknown} unknown subjects and \subref*{fig:CCR:background} false positive detections, illustrating the rate of correctly rejected samples (i.e., samples not identified as any known subject) across varying thresholds that are based on the number of correctly identified known subjects per image. 
        X-axes are plotted in logarithmic scale.
    }
\end{figure*}

While not the main purpose of this challenge, it is worth looking into closed-set performances of the algorithms, which correspond to the right-most points in \sfig{FROC:test} and \sfig{OROC:test}.
These numbers are provided in \tab{closeset}.
For the identification task, also the number of FPIPI corresponding to the highest TPIR is reported. 
While nearly all known faces are detected by most models' detectors at a rate close to 99\% -- with the exception of the DERMALOG detectors -- TPIR varies significantly across algorithms.
DERMALOG's models detect approximately 83\% of known subjects, fewer than other models, due to their facial landmark detection step that assesses each landmark's quality, effectively eliminating low-quality faces to reduce false positives during the recognition phase.
However, DERMALOG models still demonstrate commendable consistency, correctly identifying about 79\% of the faces (83\%), and they boast the lowest FPIPI of 0.84.
Among the models, AdaFace stands out by correctly identifying 92\% of the faces it detects (98\%), showcasing the best performance at the second-lowest FPIPI.
We can conclude that DERMALOG models are particularly suited for environments where minimizing false positives is important, whereas models like V2IP-AdaFace, V2IP-MEL, and UZH-DaliFace are better suited for settings where higher false identifications are acceptable due to their higher TPIR.

\subsection{Analysis of Threshold}

When evaluating detection and identification models via FROC and O-ROC, thresholds are estimated on the test set directly.
This does not correspond to operation conditions where thresholds have to be determined before deployment \cite{pereira20228years}.
To test this behavior, we determine detection thresholds $\theta_i$ in \eqref{eq:sum-tpdr} and recognition thresholds $\tau_i$ via \eqref{eq:sum-tpir} on the validation set, and compute all metrics on the test set.
In \fig{thresholds}, we show the effects of the different ways of selecting the thresholds on the final evaluation.
While for some methods, the thresholds from the validation set translate well to the results on the test set, i.e., the FPDPI and FPIPI do not change much, for other methods these numbers are less stable, resulting in a large performance difference.
This highlights the need for more realistic evaluation metrics used in face detection and open-set face recognition tasks.

\subsection{Analysis of the Unknown}

Since our test data contains two different types of unknowns, \ie, unknown faces $\mathbf U_{-1}$ and false positive detections $\mathbf U_{\mathrm{FPD}}$, we investigate the behavior of the recognition systems on both types separately.
Similar to \cite{guenther2017challenge,guenther2018challengeposter}, we plot the Correct Unknown Rejection Rate (CURR) over the True Positive Identifications Per Image (TPIPI):
\begin{gather}
    \label{eq:TPIPI}
    \begin{aligned}
        \hspace*{-.7em}\text{TPIPI}(\tau) = \frac{1}{|\mathbf{I}|}\,\Bigl| \bigl\{ F_p \in \mathbf{K} \mid \argmax_{g \in G} s(T_g, F_p) = g^*& \\
        \land \; s(T_{g^*}, F_p) \geq \tau \bigr\} \Bigr|&
    \end{aligned}\\
    \label{eq:CURR}
    \hspace*{-.7em}\text{CURR}(\tau) = \frac{1}{|\mathbf U_\bullet|} \Big| \Big\{ F_p \in \mathbf U_\bullet \mid \max_{g \in G} s(T_g, F_p) < \tau \Big\} \Big|
\end{gather}
with $\mathbf U_\bullet\in\{\mathbf U_{-1},\mathbf U_{\mathrm{FPD}}\}$.
The CURR analysis for unknown subjects $\mathbf U_{-1}$ as shown in \sfig{CCR:unknown} reveals that most algorithms are prone to assign watchlist identity label to all unknown subjects at their low thresholds $\tau$ (high TPIPI).
UZH-DaliFace, V2IP-AdaFace, and DERMALOG diverge from this trend, albeit still exhibiting very low CURR of about 2\% at their highest TPIPI.
This common issue might show potential problems with the labels of unknown subjects, which we analyze in the Supplemental Material.
Among all models, UZH-DaliFace, V2IP-AdaFace, and DERMALOG consistently outperform others, maintaining better TPIPI across every CURR level.
Similarly, they sustain a high CURR of 90\% up to a TPIPI of 1.2 but experience a sharp decline in CURR like other all models as their rejection thresholds $\tau$ are lowered further.

In the analysis of rejection of false positive detections $\mathbf U_{\mathrm{FPD}}$, UZH-DaliFace distinguishes itself by maintaining the highest CURR across all levels of TPIPI, as highlighted in \sfig{CCR:background}. 
Impressively, it manages to avoid assigning identities to all background detections, holding a CURR of 36\% even at the highest TPIPI.
Similarly, MagFace-Baseline continues with high CURR, and then sharply starts declining the CURR as the rejection threshold decreases.
V2IP-AdaFace and V2IP-MEL also perform commendably, with relatively stable CURR as TPIPI increases.
Conversely, the TR series models -- TR-EnsembET, TR-EnsembETMN, and TR-ET -- display significantly lower performance across both unknown categories, indicating that enhancements in their algorithms could be necessary.

\section{Conclusion}

We present a comprehensive evaluation of the results from participants in the watchlist challenge, focusing on the critical aspects of real-world surveillance scenarios where open-set face detection and recognition are pivotal. 
This challenge is designed to foster collaboration and establish an ideal benchmark for assessing the robustness and performance of facial recognition algorithms in surveillance settings, incorporating revised data and protocols to reflect more realistic conditions.
Detection results generally meet expectations, even in challenging cases presented by the dataset.
However, a handful of faces under extreme conditions are not detected by any model, highlighting areas for potential improvement in detection capabilities.

In terms of identification, some models excel at strict thresholds, making them suitable for applications where minimizing false positive identifications is vital. 
Conversely, other models perform exceptionally well at softer thresholds, achieving the highest TPIR. The selection of models can therefore be tailored based on system preferences and the specific security requirements of the deployment environment.
The open-set performance of the models, particularly in terms of the Correct Unknown Rejection Rate (CURR) at lower thresholds, needs improvement. 
The results highlight that all models struggle to maintain high rejection rates as the threshold for true positive identifications decreases, indicating a crucial area for future research.

The analysis indicates that models pre-trained on large-scale datasets typically surpass others, highlighting the significant impact of extensive training on model performance.
Two models that were fine-tuned to the UCCS validation set show promising capabilities.
However, due to a limited number of identities in the training data, these designs currently exhibit poorer identification performance compared to those trained on more extensive datasets.
It is anticipated that with further training on datasets containing a larger array of identities, the performances of those models increase.


\section*{Acknowledgements}
{\small
This research was supported by the Student Abroad Program of the Republic of Türkiye Ministry of National Education.
We are thankful for their financial support and dedication to fostering academic and professional growth.
}
{\small
\bibliographystyle{ieee}
\bibliography{main}
}


\end{document}